\documentclass{article}

\usepackage{arxiv}
\usepackage{subcaption}
\usepackage[utf8]{inputenc} % allow utf-8 input
\usepackage[T1]{fontenc}    % use 8-bit T1 fonts
\usepackage{hyperref}       % hyperlinks
\usepackage{url}            % simple URL typesetting
\usepackage{booktabs}       % professional-quality tables
\usepackage{amsfonts}       % blackboard math symbols
\usepackage{nicefrac}       % compact symbols for 1/2, etc.
\usepackage{microtype}      % microtypography
\usepackage{lipsum}		% Can be removed after putting your text content
\usepackage{graphicx}
\usepackage{natbib}
\usepackage{doi}
\usepackage{amsmath}

\title{bfmd: a full-match badminton dense dataset \\for dense shot captioning}

\author{ \href{https://orcid.org/0000-0002-3067-7341}{\includegraphics[scale=0.06]{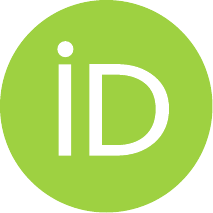}\hspace{1mm}Ning Ding}\\
	Nagoya Institute of Technology\\
	\texttt{ding.ning@nitech.ac.jp} \\
	\And
	\href{https://orcid.org/0000-0001-5487-4297}{\includegraphics[scale=0.06]{orcid.pdf}\hspace{1mm}Keisuke Fujii} \\
	Nagoya University\\
	\texttt{fujii@i.nagoya-u.ac.jp} \\
	\AND
	\href{https://orcid.org/0000-0001-9712-7777}{\includegraphics[scale=0.06]{orcid.pdf}\hspace{1mm}Toru Tamaki} \\
	Nagoya Institute of Technology\\
	\texttt{tamaki.toru@nitech.ac.jp} \\
}

\renewcommand{\shorttitle}{\textit{arXiv} Template}

\hypersetup{
pdftitle={A template for the arxiv style},
pdfsubject={q-bio.NC, q-bio.QM},
pdfauthor={David S.~Hippocampus, Elias D.~Striatum},
pdfkeywords={First keyword, Second keyword, More},
}

\begin{document}
\maketitle

\begin{abstract}
Understanding tactical dynamics in badminton requires analyzing entire matches rather than isolated clips.
However, existing badminton datasets mainly focus on short clips or task-specific annotations and rarely provide full-match data with dense multimodal annotations.
This limitation makes it difficult to generate accurate shot captions and perform match-level analysis.
To address this limitation, we introduce the first Badminton Full Match Dense (BFMD) dataset, with 19 broadcast matches (including both singles and doubles) covering over 20 hours of play, comprising 1,687 rallies and 16,751 hit events, each annotated with a shot caption.
The dataset provides hierarchical annotations including match segments, rally events, and dense rally-level multimodal annotations such as shot types, shuttle trajectories, player pose keypoints, and shot captions.
We develop a VideoMAE-based multimodal captioning framework with a Semantic Feedback mechanism that leverages shot semantics to guide caption generation and improve semantic consistency.
Experimental results demonstrate that multimodal modeling and semantic feedback improve shot caption quality over RGB-only baselines. We further showcase the potential of BFMD by analyzing the temporal evolution of tactical patterns across full matches.
	
\end{abstract}

% keywords can be removed
\keywords{Sports Video Captioning, Badminton Dataset, Sports Analysis}

\begin{figure}[t]
    \centering
    \includegraphics[width=0.4\linewidth]{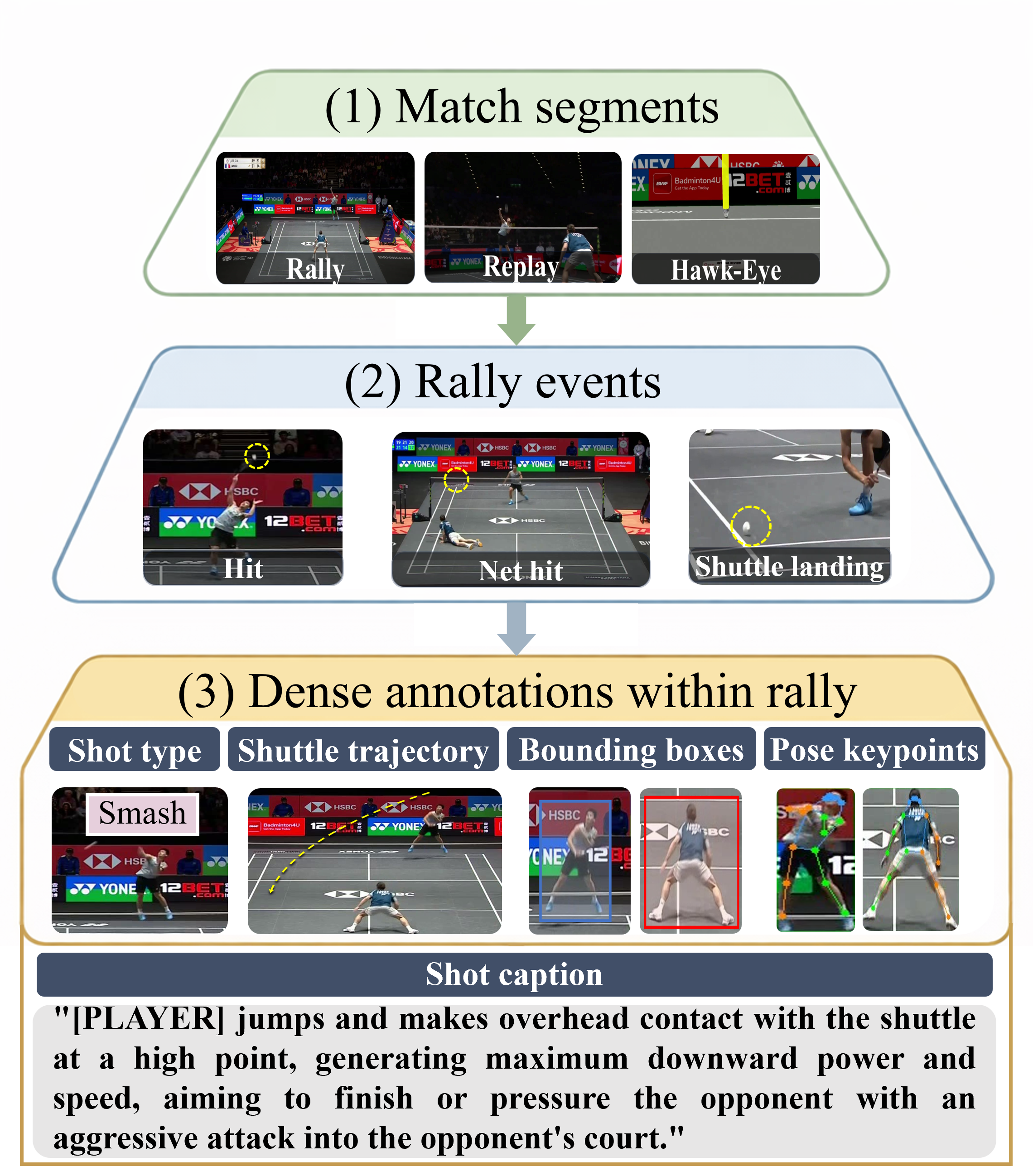}
    \caption{Example annotations from BFMD illustrating its hierarchical structure:
    (1) match segments (rallies, replays, and Hawk-Eye),
    (2) rally events (hit, net hit, shuttle landing), and
    (3) dense rally annotations including shot types, shuttle trajectories,
    player bounding boxes, pose keypoints, and shot captions.}
    \label{fig:annotation_structure}
\end{figure}

\section{Introduction}

Publicly available datasets for racket sports video understanding remain limited in both scale and structural annotation coverage. 
Existing annotations are primarily designed for specific tasks and therefore remain limited in temporal scope and modality integration.
Across various racket sports, prior works and datasets are largely task-driven, targeting problems such as event detection~\cite{voeikov2020ttnet, chang2012event, decorte2024multi}, ball tracking~\cite{sachdeva2019detection, 8909871, 9302757, chen2023tracknetv3}, shot recognition~\cite{ganser2021classification, Kulkarni_2021_CVPR, 10923883}, or player movement analysis~\cite{ding2024estimation, 10.1007/978-981-96-8298-0_11, alshami2023pose2trajectory, li2023analyzing}. 
These datasets emphasize localized objectives and typically lack annotations that capture the hierarchical structure of full broadcast matches.

Among racket sports, badminton presents additional challenges due to its rapid rally dynamics and frequent transitions between offensive and defensive states. 
The semantic meaning of each shot is often strongly conditioned on preceding rally context, making long-range temporal modeling particularly important.
In addition, accurate interpretation of badminton actions often requires complementary visual cues beyond RGB appearance. 
For example, shuttle trajectories reveal shot intent and landing patterns, while player positions and poses provide important context for understanding tactical responses and spatial interactions. 
Therefore, integrating multiple modalities is essential for generating accurate fine-grained shot descriptions.
However, existing badminton datasets are typically designed for specific tasks and provide limited multimodal or full-match supervision, focusing on tasks such as shuttle tracking \cite{8909871, 9302757, chen2023tracknetv3}, shot recognition \cite{li2024videobadminton, zhu2025analysis, mehta2024enhancing}, shot/action forecasting \cite{ijcai2024p1042, lien2025shuttleflow, chang2023will}, and shot captioning \cite{10.1145/3728423.3759408}.

Recently, FineBadminton \cite{he2025finebadminton} advanced fine-grained badminton understanding by introducing multi-level semantic annotations at the rally level. 
However, it provides limited multimodal information and is constructed from pre-segmented clips without preserving the continuous broadcast match structure.
Consequently, cross-rally dependencies and match-level dynamics remain insufficiently supported by existing datasets, highlighting the need for full-match datasets.

To address this limitation, we introduce the Badminton Full Match Dense dataset (BFMD), a match-level dataset built from full-length professional matches.
Unlike datasets constructed from pre-segmented rallies or clips, BFMD preserves the complete match timeline and provides hierarchical annotations including match segments, rally events, and dense rally-level multimodal annotations.
% This structured temporal organization enables modeling contextual dependencies both within and across rallies, supporting long-horizon match understanding beyond isolated shot analysis.
Built upon this dataset, we further investigate shot caption generation and propose a Semantic Feedback mechanism that leverages shot semantics to guide caption generation and improve semantic consistency. 
We further analyze the impact of multimodal cues, such as shuttle trajectories, player positions, and pose keypoints, on shot caption generation, as these cues explicitly capture player movement and shuttle dynamics beyond RGB appearance. 
Although BFMD is a full-match dataset, in this work we focus on shot caption generation as a first step toward match-level understanding, as reliable shot captions provide the foundation for modeling long-horizon match dynamics.

In summary, our main contributions are as follows:

\begin{itemize}
    \item We introduce BFMD dataset, the first dense full-match badminton dataset with hierarchical annotations including match segments, rally events, and dense rally-level multimodal annotations.
    
    \item We develop a VideoMAE-based multimodal captioning framework with a Semantic Feedback mechanism that leverages shot semantics to guide caption generation.
    
    \item We systematically analyze the role of multimodal cues, including shuttle trajectories, player positions, and pose keypoints, and present a qualitative analysis of tactical evolution across full matches.
\end{itemize}

\section{Related Work}

\subsection{Racket Sports Datasets}

Compared to field sports, publicly available datasets for racket sports remain limited in scale and annotation coverage.
In tennis, 3DTennisDS~\cite{skublewska2024tennis} provides a Vicon-based motion capture dataset collected from 10 professional players, while THETIS~\cite{gourgari2013thetis} contains 8,734 Kinect recordings of 12 stroke categories with RGB, depth, and skeleton data. 
In table tennis, OpenTTGames~\cite{voeikov2020ttnet} offers Full HD (120 FPS) match videos with multi-task annotations for tracking and event detection. 
Similarly, a publicly released padel dataset~\cite{decorte2024multi} includes 5.5 hours of match footage with 99 rallies and 2,377 labeled hit events.
Similar to other racket sports datasets, most existing badminton datasets
are constructed from selected rallies or clips rather than full-match broadcasts.
TrackNet~\cite{8909871} provides a badminton dataset consists of 26 broadcast videos totaling 78,200 frames and 176 annotated rallies, designed for shuttle tracking.
A drone-based badminton dataset~\cite{ding2024estimation} collects 39 doubles games with 1,347 rallies and provides shuttle locations and player bounding boxes from top-view and back-view videos.
Shot2Tactic-Caption~\cite{10.1145/3728423.3759408} consists of 10 doubles matches (approximately 7.6 hours), providing 5,494 shot captions and 544 tactic captions.
FineBadminton~\cite{he2025finebadminton} dataset built from 120 singles matches, comprising 3,215 rally clips and 33,325 shots with multi-level hierarchical annotations spanning shot types, tactical semantics, and decision evaluation.
While these badminton datasets advance fine-grained and tactical modeling, they often provide limited multimodal cues and typically annotated only at key events rather than as dense frame-level annotations. 
%
% Furthermore, only a few badminton datasets provide shot caption annotations.

\subsection{Sports Video Captioning}
Recent advances in multimodal large language models (MLLMs) and vision-language models (VLMs) have enabled natural language description and reasoning over sports videos. 
Prior work has explored video captioning and video question answering in soccer and basketball~\cite{yu2018fine, suglia2022going, qi2023goal, held2024x}, typically generating descriptions at the event level.
Beyond single-event descriptions, dense sports video captioning~\cite{mkhallati2023soccernet, rao2024matchtime} aims to generate multiple temporally localized descriptions within videos.

In badminton, recent works extend this paradigm to racket sports analysis. 
Shot2Tactic-Caption~\cite{10.1145/3728423.3759408} detects rally boundaries and shot segments, and employs a prompt-guided dual-branch captioning framework to generate both shot-level and multi-shot tactic-level descriptions from badminton videos.
%
% FineBadminton~\cite{he2025finebadminton} introduces a large-scale badminton dataset and benchmark with multi-level semantic annotations, supporting fine-grained sports video understanding and reasoning for MLLMs.
%
However, existing approaches primarily extend the temporal scope of description, while less attention has been paid to improving the semantic accuracy and completeness of shot-level captions.

\section{Dataset}

\subsection{Data Collection}

We collect full-length sports match videos from official 
BWF World Tour Super 1000 tournaments, including 
the China Open, Malaysia Open, 
All England Open, and Indonesia Open, sourced from publicly available 
broadcast recordings released by the BWF via its official YouTube channel~\cite{bwf_youtube}. 
These tournaments represent the highest competitive tier in 
international badminton, ensuring professional-level gameplay, 
consistent broadcast quality, and rich tactical dynamics.

\subsection{Event and Segment Annotation}

All temporal annotations are manually created using the Label Studio~\cite{labelstudio} with frame-level precision.
First, broadcast videos are segmented into rallies and broadcast
interruption segments, including replay segments and Hawk-Eye review
segments. 
Rally boundaries are determined based on shuttle contact and
point transitions, while replay segments are identified based on the appearance of broadcast replay overlays, and
Hawk-Eye segments are identified based on the presence of the 3D trajectory reconstruction visualizations provided by the BWF Hawk-Eye system.
Within each rally, annotators label fine-grained events including hit
events, shuttle landing events, and net hit events. Hit events correspond
to frames where a player strikes the shuttle, while shuttle landing denote
the first frame in which the shuttle visibly contacts the court surface,
and therefore occur at most once per rally.
Net hit events correspond to clear shuttle-net collisions during play.

% For each hit event, we additionally assign a shot category
% selected from predefined shot types (serve,
% long serve, smash, clear, drop, push, net shot, net kill, lift,
% drive, block, and press).

\begin{table}[t]
\centering
\caption{Dataset statistics of the proposed BFMD dataset.}
\label{tab:dataset_stats}
\begin{tabular}{lccc}
\hline
\textbf{Category} 
& \textbf{All} 
& \textbf{Singles} 
& \textbf{Doubles} \\
\hline

Matches 
& 19 
& 12 
& 7 \\

Total duration (hours) 
& 20.32 
& 13.31 
& 7.02 \\

Rallies 
& 1,687 
& 1,054 
& 633 \\

Replays 
& 795 
& 514 
& 281 \\

Hawk-Eye challenges 
& 52 
& 38 
& 14 \\

Hits
& 16,751 
& 11,301 
& 5,450 \\

Net hits
& 419 
& 210 
& 209 \\

Shuttle landings
& 1,556 
& 973 
& 583 \\

\hline
Avg. hits per rally 
& 9.93 
& 10.72 
& 8.61 \\
\hline
\end{tabular}
\end{table}

\subsection{Caption Annotation Scheme}

Each shot is represented by a hit event, corresponding to the frame where the shuttle is struck by a player. 
The hit frame serves as the temporal anchor, and surrounding frames are used for shot captioning.

To ensure semantically consistent shot descriptions, we adopt a
human-in-the-loop annotation protocol assisted by multimodal large language models.
For each shot, 16 surrounding frames (3 pre-hit and 12 post-hit) are provided to a GPT-4.1 model~\cite{achiam2023gpt}
through the API interface, as shot type is strongly correlated with
post-hit shuttle trajectory. The model is instructed to generate
structured output containing (1) a shot type selected from predefined shot types (serve, long serve, smash, clear, drop, push, net shot, net kill, lift, drive, block, and press) and (2) a short natural language description explaining how the shot is executed.
The predicted shot type is manually verified. If incorrect,
it is corrected and fed back into the prompt to regenerate
the caption. Each caption is reviewed by at least
three annotators with more than five years of badminton experience.

\begin{figure}[t]
    \centering
    
    % ---------- 第一行 ----------
    \begin{subfigure}{0.85\linewidth}
        \centering
        \includegraphics[width=\linewidth]{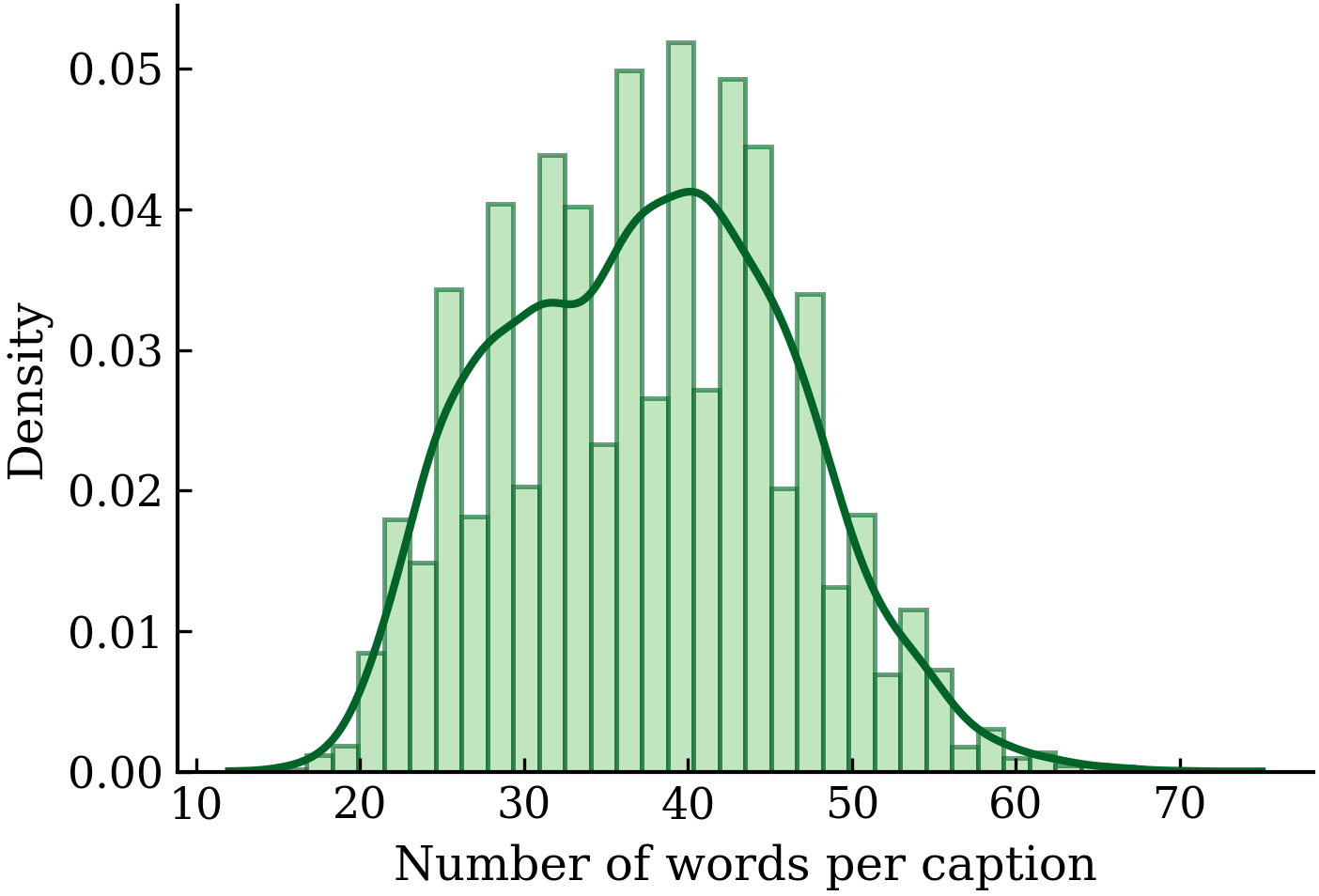}
        \caption{Distribution of caption length.}
        \label{fig:caption_length}
    \end{subfigure}
    
    \vspace{0.5em}
    
    % ---------- 第二行 ----------
    \begin{subfigure}{0.85\linewidth}
        \centering
        \includegraphics[width=\linewidth]{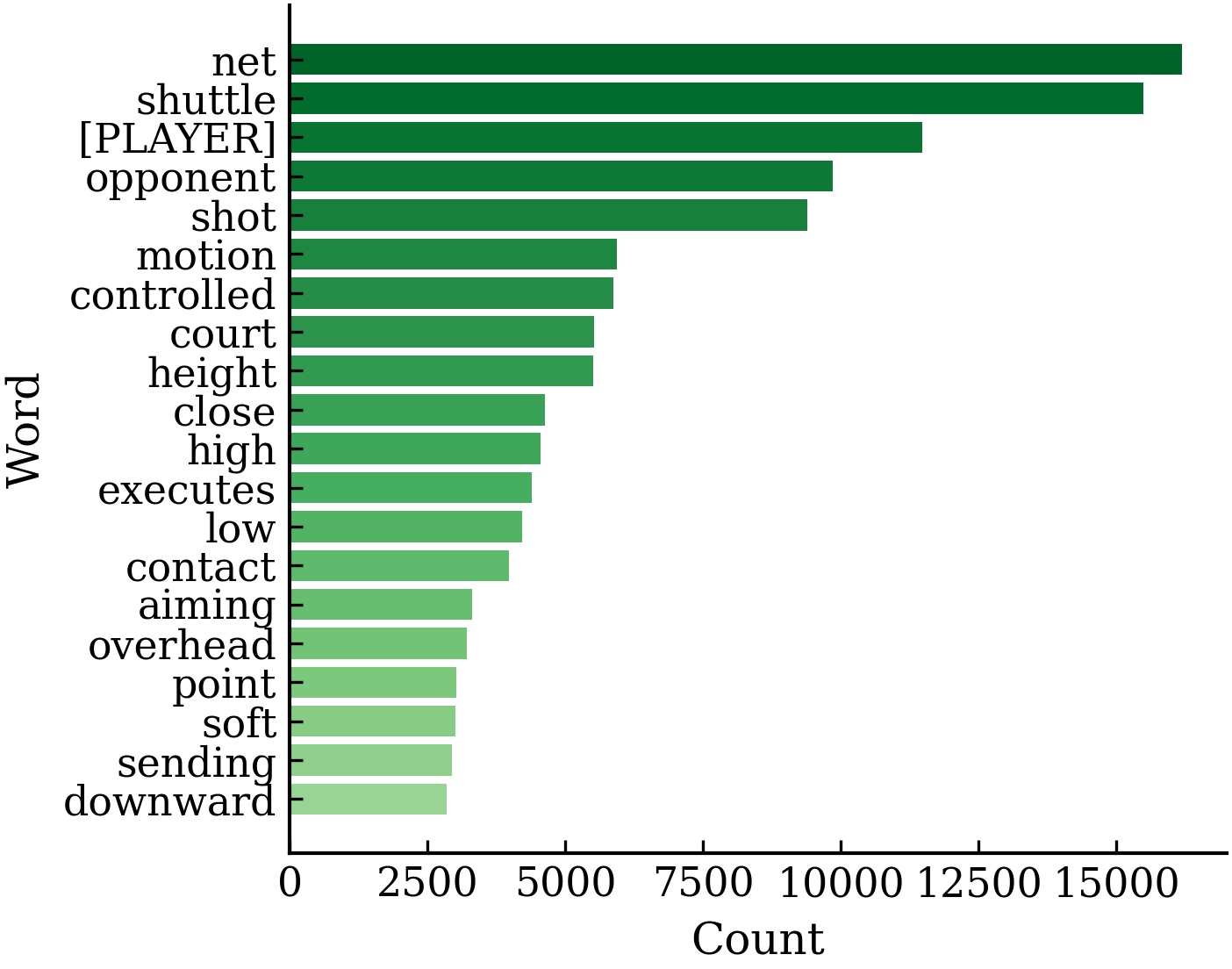}
        \caption{Top frequent words in captions.}
        \label{fig:common_words}
    \end{subfigure}
    
    \caption{Statistical analysis of shot captions.}
    \label{fig:caption_analysis}
\end{figure}

\begin{figure*}[t]
    \centering
    \includegraphics[width=0.9\textwidth]{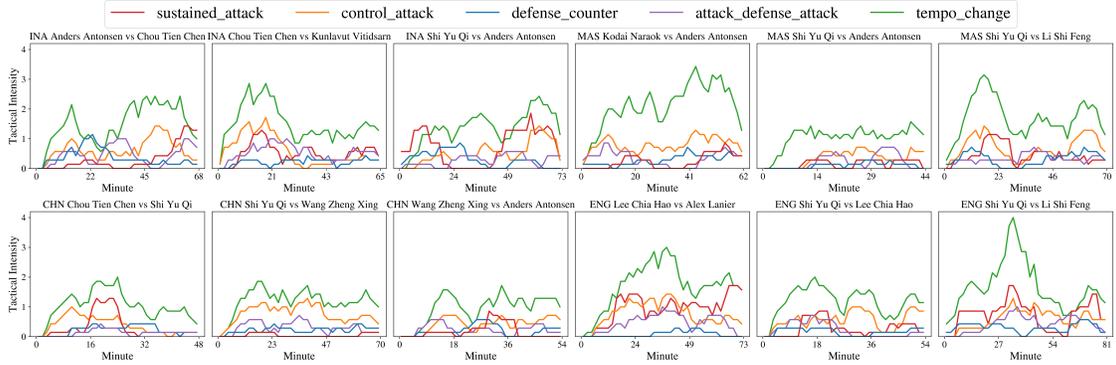}
    \caption{Tactical evolution across full-length matches. 
    Tactical patterns are detected from shot sequences and visualized over match time. 
    Each curve indicates the temporal intensity of a tactical pattern.}
    \label{fig:tactic_evolution}
\end{figure*}

\subsection{Data Statistics}

Table~\ref{tab:dataset_stats} summarizes the detailed statistics of the dataset.
Our dataset consists of 19 full-length matches, including 
12 singles matches and 7 doubles matches, 
with a total duration of 20.32 hours of broadcast footage. 
Across all matches, we annotate 1,687 rallies and 16,751 hit events.
Each hit event is associated with a corresponding caption.

Figure~\ref{fig:caption_analysis} further provides a quantitative analysis of the shot captions in dataset.
Figure~\ref{fig:caption_analysis} (a)  illustrates the distribution of caption lengths, which concentrates around 40 words, suggesting a controlled annotation style with moderate verbosity.
Figure~\ref{fig:caption_analysis} (b) shows the most frequent words, highlighting badminton-specific terminology and action-oriented verbs.
In this study, player identities are anonymized and consistently denoted as [PLAYER].

\subsection{Full-Match Tactical Analysis}

Beyond shot-level caption generation, our structured match-level annotations enable qualitative tactical analysis across entire matches. To explore the macro-level dynamics of badminton gameplay, we analyze the temporal distribution of predefined tactical patterns derived from shot sequences.
Specifically, we first map fine-grained shot types into higher-level tactical categories (e.g., attack, control, and defense). We then detect predefined tactical patterns using sliding-window matching over the categorized shot sequences. For each match, the occurrences of these patterns are aggregated over time and smoothed to visualize their temporal evolution throughout the full match duration.

Figure~\ref{fig:tactic_evolution} illustrates the tactical evolution for multiple full-length matches. Each curve represents the temporal intensity of a dominant tactical pattern. The results reveal dynamic strategic transitions across different match phases. For example, certain matches exhibit sustained attacking dominance during early stages, while others demonstrate increased defensive-counter patterns in later phases.

This analysis highlights the broader potential of our match-structured dataset for macro-level broadcast sports understanding. 
The visualization shows that structured tactical patterns naturally emerge over time, supporting future research on match-level reasoning and strategy analysis.

\section{Methodology}
\begin{figure*}[t]
    \centering
    \includegraphics[width=0.9\textwidth]{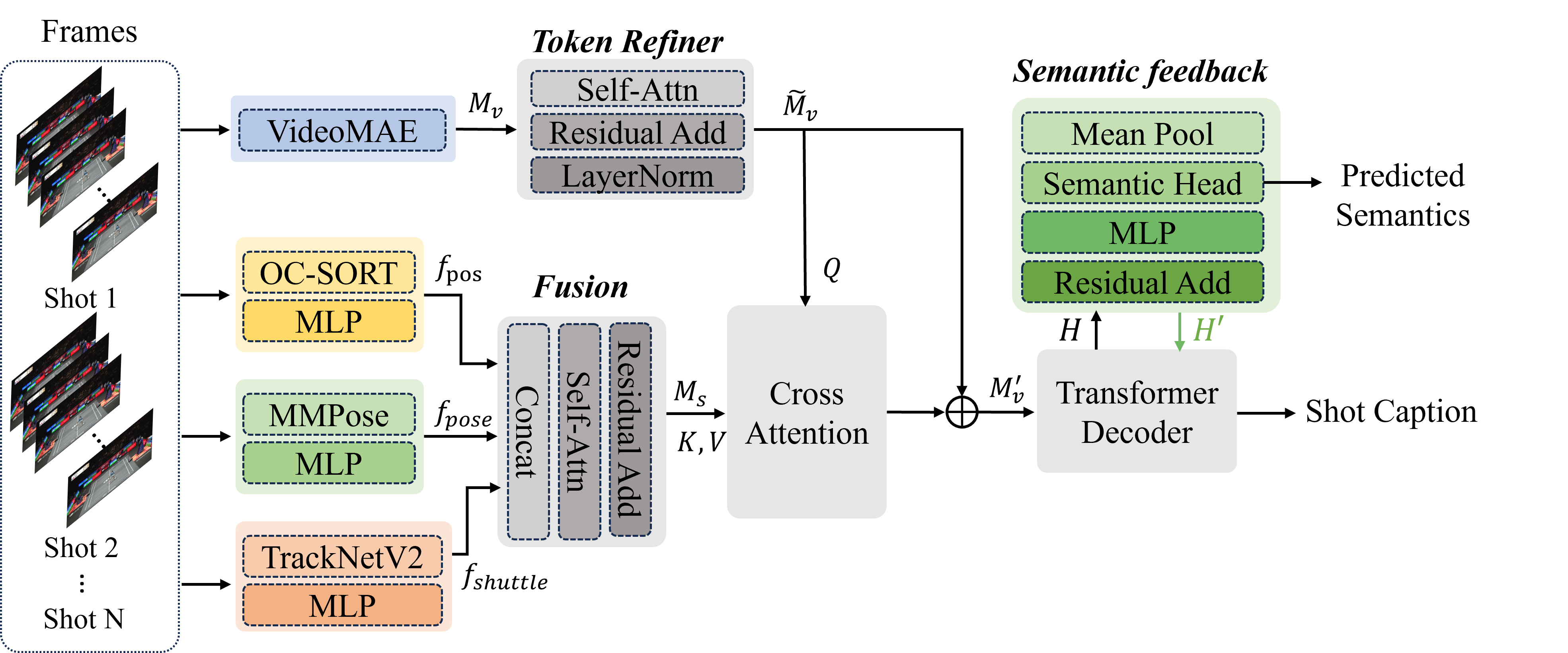}
    \caption{Overview of the proposed VideoMAE-based multimodal captioning framework with a Semantic Feedback module that leverages shot semantics to guide caption generation.}
    \label{fig:model_overview}
\end{figure*}

\subsection{Overview}

Although Transformer-based captioning models can generate fluent descriptions,
they often struggle to maintain semantic consistency when describing fine-grained sports actions.
Small visual differences between shot types may lead to incorrect or ambiguous captions.
To address this challenge, we develop a VideoMAE-based~\cite{tong2022videomae} multimodal captioning framework with Semantic Feedback (SF) that leverages shot semantics to guide caption generation and improve semantic consistency.

As illustrated in Fig.~\ref{fig:model_overview}, our model consists of four components:
(1) a VideoMAE-based visual encoder with a lightweight Token Refiner (TF) module to enhance token interactions,
(2) a multimodal fusion module integrating bounding box, pose, and shuttle cues,
(3) a Transformer-based caption decoder,
(4) the Semantic Feedback module.

\subsection{Visual Encoding}

Given an input video clip 
\(
I \in \mathbb{R}^{T \times H \times W \times 3},
\)
\noindent where \(T\) denotes the number of frames and \(H\) and \(W\) denote the frame height and width, respectively, 
we employ VideoMAE as the visual backbone:
\begin{equation}
M_v = \text{VideoMAE}(I),
\end{equation}
where \( M_v \in \mathbb{R}^{N \times D} \) denotes the patch-level visual tokens, 
\(N\) is the number of spatiotemporal tokens, and \(D\) is the embedding dimension.

VideoMAE extracts patch-level spatiotemporal tokens but does not explicitly
model token interactions. This is particularly limiting in badminton videos
where the background is largely static and subtle motion cues are critical.
Therefore, we employ a Multi-Head Self-Attention (MHSA)~\cite{vaswani2017attention} based token refiner.

Given the initial visual tokens $M_v$, the refiner first projects them into
Query ($Q$), Key ($K$), and Value ($V$) representations:
\begin{equation}
Q = M_v W^Q,\quad
K = M_v W^K,\quad
V = M_v W^V,
\end{equation}
where $W^Q, W^K, W^V \in \mathbb{R}^{D \times D}$ are learnable
projection matrices.
Each attention head is computed as
\begin{equation}
\text{head}_i =
\text{Softmax}\left(
\frac{Q_i K_i^{\top}}{\sqrt{d_k}}
\right)V_i,
\end{equation}
and the multi-head attention output is
\begin{equation}
M_v' =
\text{Concat}(\text{head}_1,\ldots,\text{head}_h)W^O, 
\end{equation}
where $h$ denotes the number of attention heads and $W^O$ is the output projection matrix.
To preserve the original spatial information and ensure numerical stability
during training, we apply a residual connection followed by Layer Normalization~\cite{ba2016layer}:
\begin{equation}
\tilde{M}_v =
\text{LayerNorm}(M_v + M_v'),
\end{equation}
where $\tilde{M}_v \in \mathbb{R}^{N \times D}$ represents the final
enhanced visual tokens. This refined representation is then passed to the
Transformer decoder to guide the caption generation.

\subsection{Multimodal Fusion}

Appearance-based visual features alone are often insufficient for
distinguishing fine-grained badminton actions. For example, smash shot and drop shot may exhibit similar visual patterns but differ
significantly in shuttle trajectory and player motion. Therefore,
we incorporate additional cues including player positions, poses,
and shuttle trajectories.

For each shot, the multimodal inputs consist of player positions
\( X = \{X_1, X_2\} \), pose keypoints
\( P = \{P_1, P_2\} \) corresponding to the two players on court,
together with the shuttle trajectory \( S \).
Player positions are estimated from the detected bounding boxes by
taking the center point of the bottom edge of each bounding box,
which approximates the players' positions on the court.
The player ordering is fixed according to broadcast layout,
ensuring consistent correspondence across frames.

Each modality is encoded using a modality-specific MLP:
\begin{align}
f_{pos} &= \text{MLP}_{pos}(X), \\
f_{pose} &= \text{MLP}_{pose}(P), \\
f_{shuttle} &= \text{MLP}_{shuttle}(S).
\end{align}

To model cross-modal interactions, we concatenate the modality
embeddings into multimodal tokens:
\begin{equation}
F_s = [f_{pos} \, \| \, f_{pose} \, \| \, f_{shuttle}],
\end{equation}
and apply multi-head self-attention:
\begin{equation}
M_s = \text{MHSA}(F_s) + F_s.
\end{equation}

Given the refined visual tokens $\tilde{M}_v$, 
we employ cross-attention to allow visual tokens to selectively
attend to relevant multimodal cues.
\begin{equation}
\Delta M_v =
\text{Attention}(\tilde{M}_v, M_s, M_s).
\end{equation}
The multimodal-enhanced visual tokens are obtained as:
\begin{equation}
M_v' = \tilde{M}_v + \alpha \Delta M_v,
\end{equation}
where $\alpha$ controls the influence of multimodal cues and we set $\alpha = 0.2$.

\subsection{Caption Decoder}

We adopt a Transformer decoder to generate captions autoregressively.
Given previously generated tokens $y_{<t}$, we first obtain their
token embeddings and positional encodings to form the decoder input.

At decoding step $t$, the decoder applies masked self-attention over
the previous tokens to produce the decoder hidden state $Q_t$.
This hidden state then attends to the multimodal-enhanced visual
tokens $M_v'$ via cross-attention:
\begin{equation}
H_t =
\text{Attention}(Q_t, M_v', M_v'),
\end{equation}
\noindent where $Q_t$ is the decoder query at step $t$, and $M_v'$
denotes the multimodal-enhanced visual tokens. The resulting hidden
representation $H_t$ is passed through a feed-forward network, followed
by a linear projection and softmax, to predict the next token.

\subsection{Semantic Feedback}

To explicitly incorporate shot-level semantics into caption generation,
we predict semantic attributes such as shot type, trajectory, and court region
from decoder hidden states and use them to refine the decoder representations.

Given decoder hidden states
$H \in \mathbb{R}^{B \times L \times D}$,
where $B$ denotes the batch size, $L$ is the caption length, and $D$ denotes the hidden dimension,
we first obtain a sentence-level representation by average pooling:
\begin{equation}
z = \frac{1}{L} \sum_{t=1}^{L} H_t,
\end{equation}
where $z \in \mathbb{R}^{B \times D}$.
We then predict semantic logits from the sentence-level representation:
\begin{equation}
S = W_s z,
\end{equation}
where $W_s \in \mathbb{R}^{K \times D}$ and $K$ denotes
the number of predefined semantic categories.
The semantic probabilities are obtained by
\begin{equation}
P = \sigma(S),
\end{equation}
\noindent
where $\sigma(\cdot)$ denotes the sigmoid function.

To incorporate semantic feedback into the decoder representations,
we project the semantic probabilities back into the hidden space
through a two-layer MLP with GELU activation:
\begin{equation}
\Delta h = W_2 \, \phi(W_1 P),
\end{equation}
where $W_1 \in \mathbb{R}^{D \times K}$,
$W_2 \in \mathbb{R}^{D \times D}$,
and $\phi(\cdot)$ denotes the GELU activation.

The decoder representations after semantic feedback are obtained as:
\begin{equation}
H' = H + \beta \cdot \Delta h,
\end{equation}
where $\beta$ is a learnable scaling parameter controlling the strength of semantic feedback, initialized to $0.1$.

\begin{table}[t]
\centering
\caption{Definition of key semantic attributes.}
\label{tab:semantic_attributes}
\setlength{\tabcolsep}{6pt}
\renewcommand{\arraystretch}{1.1}

\begin{tabular}{l p{0.45\linewidth}}
\hline
\textbf{Group} & \textbf{Attributes} \\
\hline

Shot Category (12) &
serve, long serve, smash, clear, drop, push, net shot, net kill, lift, drive, block, and press \\

Trajectory \& Intensity (4) &
high / upward / arc, downward / steep,
flat / horizontal, soft / gentle / controlled \\

Court Region (3) &
forecourt, mid-court, backcourt \\

Tactical Intent (3) &
attack / aggressive / finish,
defensive / recover / reset,
pressure / disrupt \\
\hline
\end{tabular}
\end{table}

\subsection{Training Objective}

The model is optimized using caption generation loss
and structured semantic supervision.

\paragraph{Caption Loss}

Let $\hat{Y} \in \mathbb{R}^{B \times L \times V}$ denote the predicted token logits.
where $B$ is batch size,
$L$ is sequence length,
and $V$ is vocabulary size.
We apply token-level cross-entropy loss:
\begin{equation}
\mathcal{L}_{cap}
=
\text{CrossEntropy}(\hat{Y}, y),
\end{equation}
where $y$ is the ground-truth token sequence shifted by one position,
and padding tokens are ignored.

\paragraph{Semantic Feedback Loss}

Let $S \in \mathbb{R}^{B \times K}$
denote the predicted semantic logits obtained from the
sentence-level decoder representation,
where $B$ is the batch size and $K$ is the number of semantic attributes.
Let $S^* \in \{0,1\}^{B \times K}$
denote the corresponding ground-truth semantic vectors.

We apply multi-label binary cross-entropy loss:
\begin{equation}
\mathcal{L}_{sf}
=
\text{BCEWithLogits}(S, S^*).
\end{equation}

\paragraph{Total Loss}

The overall objective is:
\begin{equation}
\mathcal{L}_{total}
=
\mathcal{L}_{cap}
+
\lambda \mathcal{L}_{sf},
\end{equation}
where $\lambda$ balances caption generation
and semantic supervision.
In our experiments, we set $\lambda = 0.1$.

\begin{table*}[t]
\centering
\caption{Comparison with existing captioning approaches using RGB-only visual inputs.}
\label{tab:comparison}
\begin{tabular}{lccccccc}
\hline
Model 
& BLEU-1 
& BLEU-2 
& BLEU-3 
& BLEU-4 
& METEOR 
& ROUGE-L 
& CIDEr \\
\hline

\multicolumn{8}{c}{Vision-based Sports Captioning Models} \\
\hline

SoccerNet-Caption
& 38.4 & 24.8 & 16.2 & 10.7 
& 20.1 & 32.8 & 11.9 \\

Shot2Tactic
& 45.0 & 30.1 & 20.8 & 14.6 
& 22.8 & 34.9 & 27.9 \\

% FineBadminton
% & - & - & - & - 
% & - & - & - \\

\hline
\multicolumn{8}{c}{Pretrained Video-Language Models} \\
\hline

Vid2Seq 
& 41.2 & 25.1 & 16.4 & 11.5 
& 23.5 & 31.5 & 21.5 \\

InternVideo2 
& 42.4 & 27.4 & 18.4 & 12.7 
& 22.8 & 33.4 & 23.0 \\

\hline
\multicolumn{8}{c}{Large Vision-Language Models (Zero-shot)} \\
\hline

Qwen2.5-VL-7B-Instruct
& 18.8 & 6.5 & 2.2 & 1.0 
& 14.6 & 15.5 & 2.2 \\

Qwen3-VL-8B-Instruct
& 17.2 & 8.2 & 3.0 & 1.3
& 14.8 & 16.7 & 2.4 \\

GPT-4.1 
& 22.0 & 12.9 & 7.8 & 5.2 
& 16.9 & 23.0 & 6.8 \\

GPT-5.2
& 24.2 & 12.4 & 6.4 & 3.7 
& 16.2 & 21.4 & 6.0 \\

\hline
\textbf{Ours} 

% & \textbf{46.0} 
% & \textbf{30.8} 
% & \textbf{21.5} 
% & \textbf{15.3} 
% & \textbf{23.8} 
% & \textbf{35.5} 
% & \textbf{29.4} \\

% & \textbf{46.7} 
% & \textbf{31.0} 
% & \textbf{21.7} 
% & \textbf{15.6} 
% & \textbf{23.8} 
% & \textbf{35.7} 
% & \textbf{30.3} \\

& \textbf{47.1} 
& \textbf{31.5} 
& \textbf{21.9} 
& \textbf{15.7} 
& \textbf{23.7} 
& \textbf{35.9} 
& \textbf{32.3} \\

\hline
\end{tabular}
\end{table*}

% \begin{table}[t]
% \centering
% \caption{Ablation study on architectural components.
% B4: BLEU-4, M: METEOR, R-L: ROUGE-L, C: CIDEr.}
% \label{tab:ablation_arch}
% \begin{tabular}{lcccc}
% \hline
% \textbf{Model} 
% & B4 
% & M 
% & R-L 
% & C \\
% \hline

% Baseline
% & 14.5 & 22.9 & 35.1 & 26.3 \\

% + Refiner 
% & 15.3 
% & \textbf{23.8} 
% & 35.5 & 29.4 \\

% + semantic feedback 
% & 15.6 & 23.3 & 35.6 & 27.7 \\

% Refiner + Semantic Feedback
% & \textbf{15.7} 
% & 23.7 
% & \textbf{35.9} 
% & \textbf{32.3} \\

% \hline
% \end{tabular}
% \end{table}

\begin{table}[!t]
\centering
\caption{Ablation study on architectural components.
TR: Token Refiner. SF: Semantic Feedback.
B4: BLEU-4, M: METEOR, R-L: ROUGE-L, C: CIDEr.}
\label{tab:ablation_arch}
\begin{tabular}{cc|cccc}
\hline
\textbf{TR} & \textbf{SF}
& B4 & M & R-L & C \\
\hline

 &  & 14.5 & 22.9 & 35.1 & 26.3 \\

\checkmark &  & 15.3 & \textbf{23.8} & 35.5 & 29.4 \\

 & \checkmark & 15.6 & 23.3 & 35.6 & 27.7 \\

\checkmark & \checkmark & \textbf{15.7} & 23.7 & \textbf{35.9} & \textbf{32.3} \\

\hline
\end{tabular}
\end{table}

\section{Experiments}
\subsection{Experimental Setup}

All experiments are conducted on the singles subset of the BFMD dataset, consisting of 12 matches. 
We focus on singles to maintain a consistent two-player scenario, as doubles involve four players and varying numbers of multimodal inputs.
Also, we focus on caption generation and do not perform event detection. 
Shot events are provided by ground-truth annotations, and the corresponding frames are used as inputs.
For caption generation, each sample corresponds to a shot,
with 16 surrounding frames (3 pre-hit and 12 post-hit).
The dataset is split into training, validation, and test sets
with a ratio of 70\%, 20\%, 10\%.
We evaluate caption quality using evaluation metrics,
including BLEU~\cite{10.3115/1073083.1073135}, METEOR~\cite{banerjee2005meteor}, ROUGE-L~\cite{lin2004rouge}, and CIDEr~\cite{Vedantam_2015_CVPR}.

\subsection{Implementation Details}

We use VideoMAE-base as the visual backbone.
Input clips consist of 16 frames resized to
\(224 \times 224\) resolution.
The patch-level visual features are refined using a lightweight
Token Refiner, implemented as a single multi-head self-attention
layer with 8 attention heads followed by residual connection
and Layer Normalization.

Player bounding boxes are detected using a YOLOX detector~\cite{yolox2021}
and tracked across frames using OC-SORT~\cite{cao2023observation}.
The resulting bounding boxes are used as inputs for a top-down human pose estimation model implemented in the MMPose framework~\cite{mmpose2020}.
Shuttle trajectories are extracted using TrackNetV2~\cite{9302757}.
All structural modalities are generated automatically in a preprocessing
stage and remain fixed during caption training.
The structural modalities (bounding boxes, pose keypoints,
and shuttle trajectory) are projected into the same embedding
space using two-layer MLPs.
The caption decoder is a 6-layer Transformer decoder
with 8 attention heads per layer.
The maximum caption length is set to 120 tokens.
During training, we freeze all VideoMAE parameters
except for the last two Transformer blocks,
which are fine-tuned to adapt to the badminton domain.
Models are trained using the AdamW optimizer~\cite{loshchilov2017decoupled}
with an initial learning rate of \(1\times10^{-4}\)
and a batch size of 16.
Training is conducted for 30 epochs,
and the best result is selected based on the validation loss.

\begin{table*}[t]
\centering
\caption{Ablation study on multimodal inputs for shot captioning.}
\label{tab:ablation_modalities}
\begin{tabular}{lcccccccccc}
\hline
\textbf{Model} 
& BBox 
& Pose 
& Shuttle 
& BLEU-1 
& BLEU-2 
& BLEU-3 
& BLEU-4 
& METEOR 
& ROUGE-L 
& CIDEr \\
\hline
RGB only 
&  &  &  
& 47.1 & 31.5 & 21.9 & 15.7 
& 23.7 & 35.9 & 32.3 \\

+ BBox 
& \checkmark &  &  
& 47.7 & 32.0 & 22.2 & 16.0 
& 24.1 & 36.6 & 33.0 \\

+ Pose 
&  & \checkmark &  
& 47.1 & 31.3 & 21.6 & 15.5 
& 23.3 & 36.0 & 33.5 \\

+ Shuttle 
&  &  & \checkmark 
& 47.9 & 32.3 & 22.7 & 16.5 
& 24.3 & 38.2 & 35.4 \\

Full Model 
& \checkmark & \checkmark & \checkmark 
& \textbf{48.1} 
& \textbf{32.5} 
& \textbf{23.2} 
& \textbf{16.9} 
& \textbf{24.5} 
& \textbf{38.4} 
& \textbf{36.8} \\

\hline
\end{tabular}
\end{table*}

\subsection{Comparison with Existing Methods}

Table~\ref{tab:comparison} compares our method with representative vision-based captioning models, pretrained video-language models, and large vision-language models evaluated in a zero-shot manner.

Among vision-based models, our approach significantly outperforms both the SoccerNet-Caption~\cite{mkhallati2023soccernet}, Shot2Tactic~\cite{10.1145/3728423.3759408} across all evaluation metrics, demonstrating the effectiveness of structured multimodal representations.

Compared to pretrained video-language models such as Vid2Seq~\cite{yang2023vid2seq} and InternVideo2~\cite{wang2024internvideo2}, our method achieves consistent gains, particularly on higher-order metrics such as BLEU-4 and CIDEr, indicating improved long-form coherence and semantic relevance.
Notably, while large vision-language models (e.g., Qwen2.5-VL~\cite{Qwen2.5-VL}, Qwen3-VL~\cite{Qwen3-VL} and GPT variants~\cite{achiam2023gpt}) exhibit strong zero-shot performance, our method with multimodal integration enable further improvements. The proposed full model achieves the best overall performance, demonstrating the benefit of multimodal cues for shot caption generation.

\subsection{Component Ablation}

We further analyze the contribution of key architectural components,
including the Token Refiner and the Semantic Feedback Module.
Results are summarized in Table~\ref{tab:ablation_arch}.

Starting from the baseline model, introducing the Token Refiner
improves performance across evaluation metrics, indicating that refining patch-level visual tokens helps capture
spatiotemporal dynamics for more accurate shot descriptions.
Adding the semantic feedback module also improves performance
over the baseline, achieving higher BLEU-4 and ROUGE-L scores,
which indicates better semantic alignment between visual dynamics
and generated captions.
Finally, the full model that integrates both components achieves
the best overall performance across most metrics.
These results suggest that TR and SF module benefit shot captioning.

\begin{figure}[!t]
    \centering
    \includegraphics[width=0.46\textwidth]{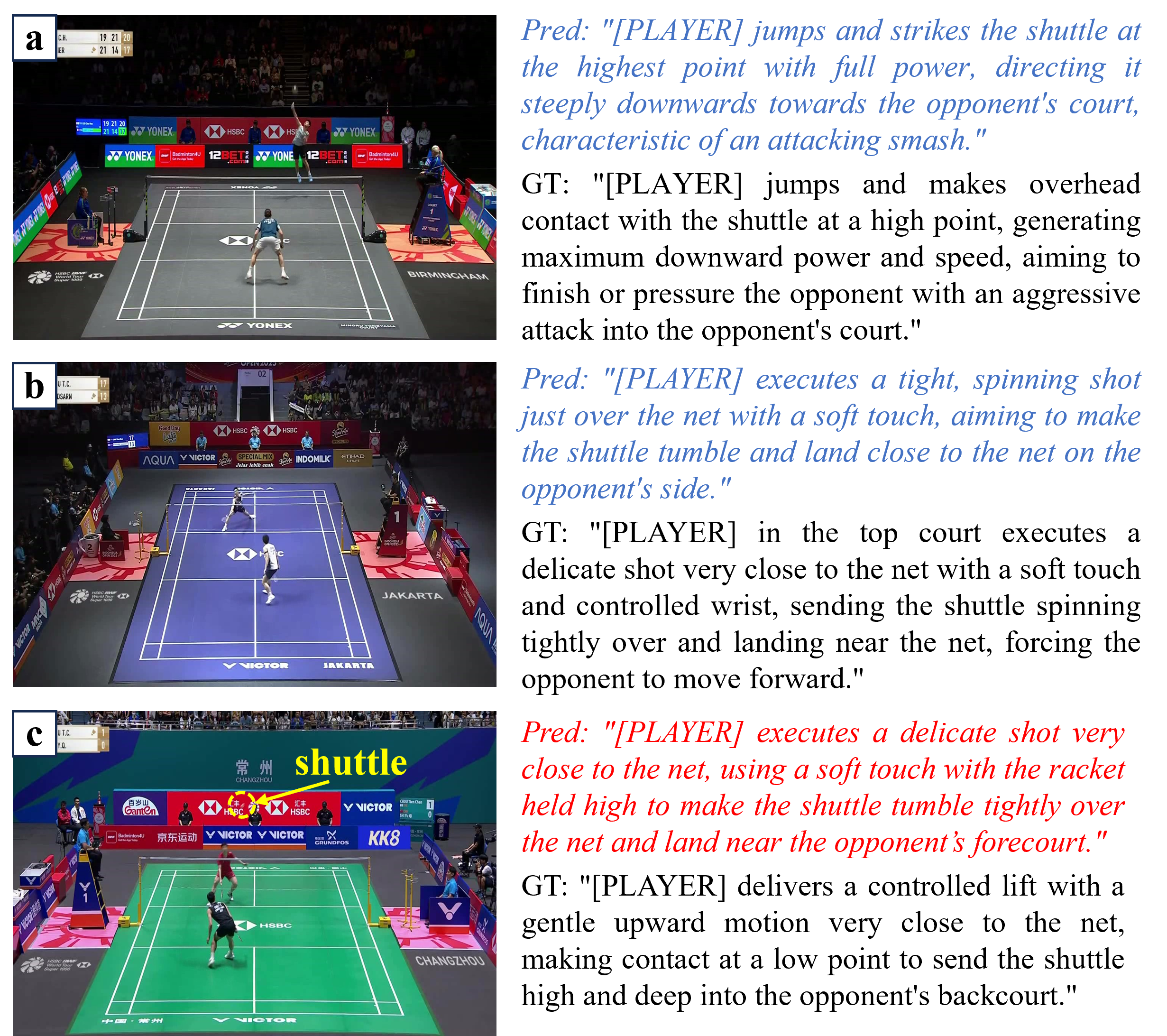}
\caption{Qualitative examples of shot captioning. 
(a)--(b) Successful predictions for smash and net shot. 
(c) A failure case where the model predicts a net shot instead of a lift due to their visual similarity.}
    \label{fig:caption}
\end{figure}

\subsection{Ablation Study on Multimodal Inputs}

Table~\ref{tab:ablation_modalities} presents an ablation study analyzing the contribution of different multimodal inputs for shot captioning. 
Starting from the RGB-only baseline, we progressively incorporate player bounding boxes, pose keypoints, and shuttle trajectory information.
Adding player bounding boxes consistently improves performance across evaluation metrics, suggesting that spatial localization provides useful structural cues beyond raw RGB features.
In contrast, incorporating pose features alone leads to marginal changes in surface n-gram metrics, with BLEU-4 and METEOR slightly decreasing. 
However, CIDEr improves, suggesting that pose information enhances higher-level semantic alignment despite limited gains in exact word overlap. 
This result suggests that pose features mainly capture fine-grained action semantics rather than directly affecting lexical patterns.
The addition of shuttle trajectory yields the most noticeable improvement among individual modalities, highlighting the importance of modeling shuttle dynamics when describing badminton shots.

Finally, the full model that integrates all modalities achieves the best overall performance. These results suggest that multimodal cues provide complementary information for shot caption generation.

\subsection{Qualitative Results}

Fig.~\ref{fig:caption} illustrates representative captioning examples, including both successful predictions and a typical failure case.
Fig.~\ref{fig:caption} (a) and (b) show examples where the model accurately captures key semantic components of the rally, including shot type and tactical intent. 
It correctly identifies an attacking smash with steep downward trajectory and a tight spinning net shot characterized by soft touch near the net. 
These cases demonstrate the model's ability to jointly reason over visual dynamics and structural cues such as player position and shuttle motion.

Fig.~\ref{fig:caption} (c) shows a representative failure case.
Although the ground truth corresponds to a controlled lift, the model predicts a delicate net shot. 
This error may be related to the visual similarity between these actions, as both occur near the net and involve relatively gentle shuttle contact. In this example, the shuttle motion appears relatively slow, and the limited
observation window of 12 frames after the hit may make it difficult to fully capture the trajectory.
Even with multimodal inputs including shuttle cues, such limited temporal context can still lead to ambiguity between similar forecourt shots.
Overall, most errors remain semantically close to the ground truth rather than entirely unrelated, indicating that the model captures general rally context but still struggles with fine-grained shot discrimination.

\section{Conclusion}

In this work, we introduced BFMD, a full-match badminton dataset that preserves complete match structures and provides hierarchical annotations including rallies, hit events, and other dense rally annotations.
We further proposed a multimodal shot captioning framework with semantic feedback that integrates player position, pose, and shuttle trajectory information. Experimental results demonstrate that multimodal cues and semantic feedback improve caption quality over RGB-only and pretrained baselines.
In future work, we aim to extend our framework toward full match video understanding, enabling temporally coherent modeling of tactical evolution and match-level dynamics.

\bibliographystyle{unsrtnat}
\bibliography{references}  %%% Uncomment this line and comment out the ``thebibliography'' section below to use the external .bib file (using bibtex) .





% \end{thebibliography}

\end{document}